\documentclass[a4paper, conference]{IEEEtran}
\IEEEoverridecommandlockouts
\usepackage{cite}
\usepackage{amsmath,amssymb,amsfonts}
\usepackage{graphicx}
\usepackage{textcomp}
\usepackage{xcolor}
\def\BibTeX{{\rm B\kern-.05em{\sc i\kern-.025em b}\kern-.08em
    T\kern-.1667em\lower.7ex\hbox{E}\kern-.125emX}}

\usepackage{mathptmx}
\usepackage{algorithm}
\usepackage{algpseudocode}
\usepackage{float}
\usepackage{physics}
\usepackage{tabularx}
\usepackage{array}
\usepackage{multirow}
\usepackage{booktabs}
\ifCLASSOPTIONcompsoc
\usepackage[caption=false,font=normalsize,labelfont=sf,textfont=sf]{subfig}
\else
\usepackage[caption=false,font=footnotesize]{subfig}
\fi

\newcolumntype{L}[1]{>{\raggedright\arraybackslash}m{#1}}
\newcolumntype{C}[1]{>{\centering\arraybackslash}m{#1}}
\newcolumntype{R}[1]{>{\raggedleft\arraybackslash}m{#1}}

\begin{document}

\title{An Enhanced Sampling-Based Method With Modified Next-Best View Strategy For 2D Autonomous Robot Exploration\\
% {\footnotesize \textsuperscript{*}Note: }
\thanks{All authors are with Faculty of Electronics and Telecommunications, VNU University of Engineering and Technology, Hanoi, Vietnam.\tt~dongtran.robotics@gmail.com, anhph@vnu.edu.vn,  hieu.dv@vnu.edu.vn, duongtanrb@gmail.com, tungbt@vnu.edu.vn, vanntt@vnu.edu.vn}
}

\author{\IEEEauthorblockN{Dong Huu Quoc Tran, Hoang-Anh Phan, Hieu Dang Van,\\ Tan Van Duong, Tung Thanh Bui and Van Nguyen Thi Thanh}
}

% \IEEEpubid{\makebox[\columnwidth]{979-8-3503-0050-5/23/\$31.00~\copyright2023~IEEE\hfill}\hspace{\columnsep}\makebox[\columnwidth]{\hfill}}

\maketitle
% \IEEEpubidadjcol

\begin{abstract}
Autonomous exploration is a new technology in the field of robotics that has found widespread application due to its objective to help robots independently localize, scan maps, and navigate any terrain without human control. Up to present, the sampling-based exploration strategies have been the most effective for aerial and ground vehicles equipped with depth sensors producing three-dimensional point clouds. Those methods utilize the sampling task to choose random points or make samples based on Rapidly-exploring Random Trees (RRT). Then, they decide on frontiers or Next Best Views (NBV) with useful volumetric information. However, most state-of-the-art sampling-based methodology is challenging to implement in two-dimensional robots due to the lack of environmental knowledge, thus resulting in a bad volumetric gain for evaluating random destinations. This study proposed an enhanced sampling-based solution for indoor robot exploration to decide Next Best View (NBV) in 2D environments. Our method makes RRT until have the endpoints as frontiers and evaluates those with the enhanced utility function. The volumetric information obtained from environments was estimated using non-uniform distribution to determine cells that are occupied and have an uncertain probability. Compared to the sampling-based Frontier Detection and Receding Horizon NBV approaches, the methodology executed performed better in Gazebo platform-simulated environments, achieving a significantly larger explored area, with the average distance and time traveled being reduced. Moreover, the operated proposed method on an author-built 2D robot exploring the entire natural environment confirms that the method is effective and applicable in real-world scenarios.
\end{abstract}


\section{Introduction}
In robotics research fields, following the discoveries of control theories, for a mobile robot system to drive intelligently, it is necessary to utilize the three fundamental robot control theory approaches: mapping, localization, and navigation \cite{b1}. As delineated in \cite{siegwart2011introduction}, an autonomous robot must construct a model of its surrounding environment by integrating localization and mapping and facilitating safe navigation.

The advantage of this combination lies in its capacity to optimally discover and generate maps by employing specialized planners, such as goal planners, path planners, or motion planners, which perform adaptive motion and real-time decision-making for map exploration or deliberate movement. Planners that autonomously cover the map tend to belong to a category known as exploration. Subsequently, the primary tier in exploration approaches involves generating a set of feasible actions the robot could execute, such as goals.

Hitherto, Frontier Detection and Random Exploration have emerged as the most effective methodologies for goal planners \cite{placed2023survey}. These two procedures separated common exploratory planning strategies into frontier-based and sampling-based \cite{b6}. Frontier-based approaches determine their planning actions from the boundaries between free and known space, referred to as frontiers. While sampling-based strategies generally aim to select random points or develop Rapidly-exploring Random Trees (RRTs) to calculate the exploration path. Notably, the sampling-based method choosing the Next Best View (NBV) was initially introduced \cite{b16}, demonstrating its advantages in dynamic or uncertain contexts where pathways cannot be dependably precomputed. Conversely, frontier-based planners prove effective when robots possess environmental references and knowledge, allowing for expedited exploration in two-dimensional or three-dimensional spaces.

Nevertheless, in a 2D environment, the Frontier-based method occasionally leads to the robot becoming entrapped in detrimental situations due to frequent deficiencies in environmental knowledge, such as encountering uncertain obstructions and being unable to advance significantly toward its objectives. In contrast, the sampling-based NBV planner prevents the robot from entering newly discovered areas owing to its inability to circumvent the local minimum. As described in \cite{b9}, the performance of sampling-based planners deteriorates in expansive environments or constrained scenarios characterized by narrow openings or bottlenecks. This results in a more substantial time computation requirement when the robot attempts to determine the optimal goal, owing to the revisiting of previously traversed sections or irregular resultant movement, thereby causing the robot to consume a longer duration and distance to achieve a covered map.

In this paper, a method for simultaneously exploring and mapping an unknown 2D space is developed. The environment was mapped using a laser sensor-generated occupancy grid map and the NBV strategies for determining the robot's movement path. Using the RRT algorithm theory, the proposed NBV method searched for exploration paths and decided on the first destination as the NBV point. The branch's nodes were randomly selected according to a uniform distribution. Unless the number of iterations exceeds the initial setup, these branches stopped at unknown and recognized map borders. Then, our reward equation was applied to RRT vertices, decreasing computing costs and optimizing the predetermined objectives. Conditions for evaluating frontiers include the distance between two nodes and the information gain. This method achieved concentrated exploration similar to frontier-based methods, requiring fewer candidate locations for random trees and being biased to grow towards regions where the robot had not yet traveled. It aided robots in avoiding local minimums and reduced processing costs compared to RH-NBV  Sampling-based approaches.

\section{Related Work}
The study of robotics in the twenty-first century has advanced significantly, leading to consistent enhancements in robotic intelligence. Simultaneous localization and mapping (SLAM) is a collection of techniques designed to address the challenges of mapping and localization concurrently \cite{b10}. The term active SLAM was first coined by Davison \cite{davison2002simultaneous}, wherein SLAM would be integrated with active perception to control robots and reduce the uncertainty of their localization and map representation. As a result, studying active perception (also called exploration) to find optimal actions for robots' ASLAM technique is essential, eliminating the need for human control over robot movement. The first exploration using adaptive planners was introduced by Thrun et al. \cite{thrun1991active}.

With exploration tasks in mind, \cite{b2} introduced planners that choose actions and maximize knowledge of two variables of interest. This led to a new research direction: Investigating unknown areas of the environment that robots need to navigate and make evaluated decisions based on utility computation. Notably, strategies based on the NBV \cite{b12} and Frontier \cite{b13} theories are popular. The first frontier exploration research selected the closest frontier to the robot. Umari and Mukhopadhyay demonstrated the first use of the Sampling-based method with Frontier Detection, employing RRT algorithms to find frontiers \cite{b15}. Quinn et al. developed several geometric frontier-detection methods to improve the performance of previous algorithms by evaluating only a subset of observed free space \cite{quin2021approaches}. The sample-based frontier detector algorithm introduced by \cite{sun2022ada} reduces the computational load of sampling to find frontiers by sensing the surrounding environment structure and using non-uniform distributed sampling adjacent sliding windows. Soni et al. presented a novel frontier tree approach for multi-robot systems \cite{soni2022multi}.

Concerning the NBV hypothesis, the most prevalent approach is the sampling method employing rapidly random trees to determine optimal paths in known space, referred to as Receding Horizon NBV (RH-NBV) \cite{b16}. Extending this work, Bircher et al. presented another sampling-based receding horizon path planning paradigm \cite{b17}, and Papachristos et al. delivered an uncertainty-aware exploration and mapping planning strategy using a belief space-based approach \cite{b18}. To facilitate the evaluation of the NBV path cost, Wang et al. propose a graph-structured roadmap \cite{wang2019autonomous}, while Batinovic introduced a cuboid-based evaluation method that results in an enviably short computation time \cite{batinovic2022shadowcasting}.

In contrast, due to the advantages of Receding Horizon NBV and classic frontier exploration planning, Selin et al. proposed combining both techniques \cite{b19}. Dai et al. suggested a hybrid exploration approach based on sampling-based and frontier-based methods by sampling candidate next-views from the map's frontiers \cite{b20}. Lu et al. presented the Frontier enhanced NBV method using a frontier planner as a global and NBV as a local planner \cite{lu2022optimal}.

With this research, numerous studies have utilized different algorithms for routing and embedding with SLAM. Trivun et al. developed an ASLAM with Fast-SLAM, Particle Filter, A* Global Planner, and DWA Local Planner for autonomous exploration and mapping in a dynamic indoor environment \cite{b23}. Bonetto et al. developed an omnidirectional robot to find frontiers and adjust its heading using a rotation sensor while actively navigating, aiming for the robot consistently to achieve the highest features of map representation \cite{bonetto2022irotate}.

\section{Proposed Approach}
\begin{figure}[htbp]
    \centering
    \includegraphics[width=.55\columnwidth]{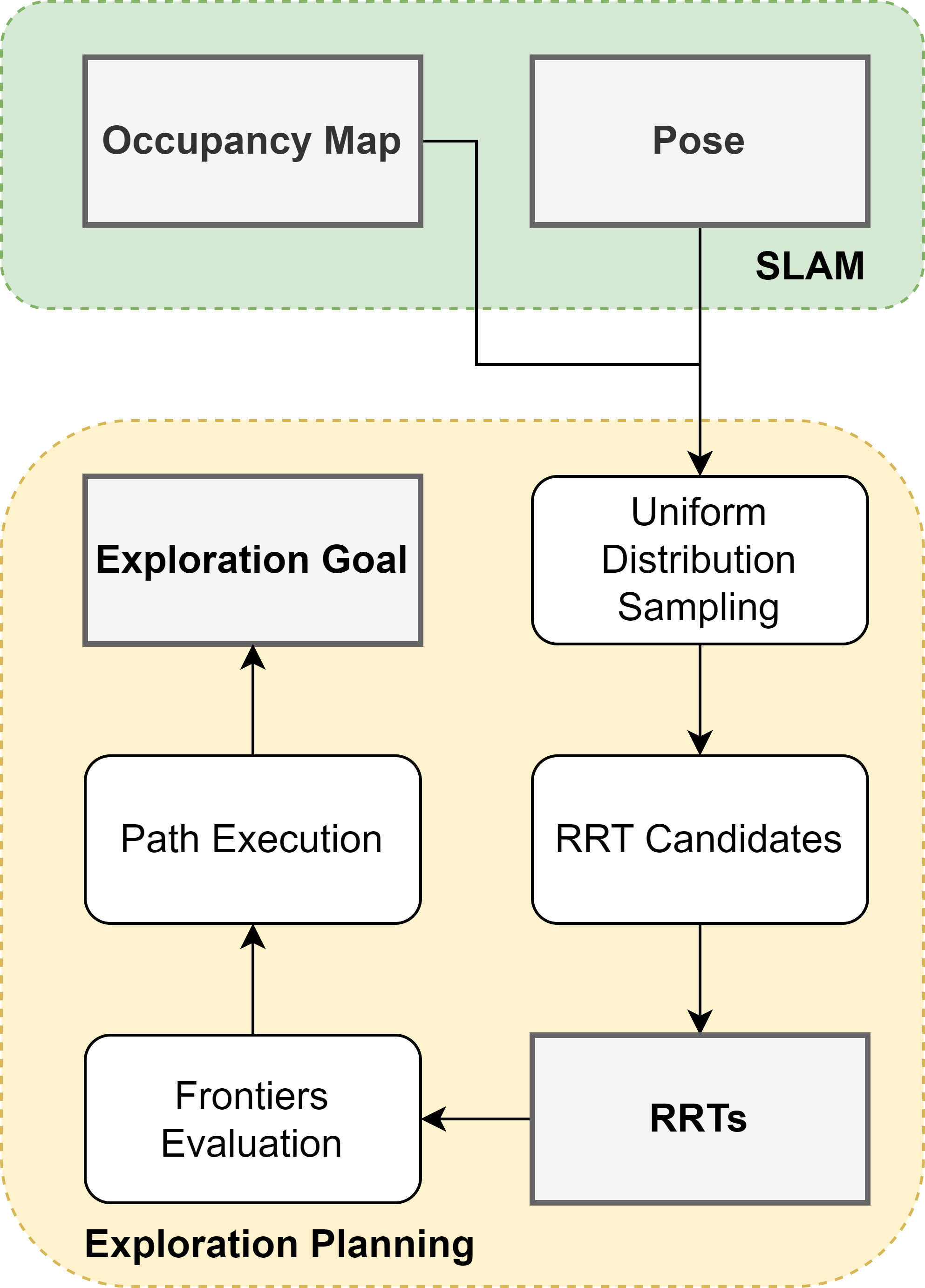}
    \caption{Diagram illustrating the proposed autonomous exploration and mapping methodology. The SLAM module runs persistently in the background as our suggested planner identifies new destinations utilizing distribution sampling and information gain-based utility measurements.}
    \label{fig:diagram}
\end{figure}

The core idea of our exploration planner remains rooted in the NBV Sampling-based Planner theory: Determine a destination and reach it by sampling RRTs, evaluate the most suitable RRT for exploration, and select the first node of the RRT as the NBV. Fig. \ref{fig:diagram} visually represents the planner's structure.%

Due to the robot's mobility and SLAM running continuously in the background, the robot's positions and scanned maps are constantly updated. A sampling-based method is employed to identify nodes as worthwhile destinations. The RRT algorithm is used as the sample-based and frontiers-based planner to find nodes or frontiers. A utility equation describing the expected information gain over time evaluates the nodes. Subsequently, the RRT vertex with the highest gain in the final node is selected. A navigation planner drives the robot to each destination from its current position to the NBV point of the chosen vertex.

Assume we have an occupancy grid map $M \subset \mathbb{R}^2$ as the total space environment set covered by sensors, with cells $\mathbf{m} \in M$ representing a two-dimensional point $\vb*{x}_M$ on a coordinate system, which has $P_o^n(\mathbf{m})$ occupancy probabilities at time $n$. This set, as the map, is frequently updated by adding new cells that the robot observes with $P_o^0(\mathbf{m})$. All previously added cells are updated using Bayesian Theory, with distribution $p(\mathbf{m} | \operatorname{occ})$, where $\operatorname{occ}$ denotes the likelihood of obstruction.

At the beginning of each planning iteration, the robot assumes a localized position and orientation, forming the two-dimensional vehicle configuration state, $x_0=(x, y, \psi)^T$. We determine the cut-off steps as $N_{max}$. In each planner sampling iteration, $N_T$ increases by $1$, and the planner terminates if $N_T$ reaches $N_{max}$. However, if the best gain $g_{best}$ remains zero, the sampling loop continues until the final node of RRT is a frontier, or $g_{best}>0$.

Predefined variables such as $\epsilon$ for tree length and $\alpha$ for overshoot view are included. A filter is applied to the RRT algorithm to adjust and eliminate uncertain nodes, dead locations, and out-of-map points. Our proposed approach steps can be outlined in Algorithm \ref{alg:hsf}.%
\begin{algorithm}
	\caption{Exploration Planner}\label{alg:hsf}
	\begin{algorithmic}[1]
	    \State $\epsilon, \alpha, N_{max} \gets initVariables()$
		\State $N_{T} \gets 0$
		\State $\vb*{x}_{best}, g_{best} \gets \vb*{x}_{0}, g_{0}$
		\State $T_{0} \gets (\vb*{x}_{r}^{0}, g_{0})$
		\While {$N_{T} < N_{max}$ \textbf{ or } $g_{best} = 0$}
			\State $\vb*{x}_{rand} \gets randomModel$
			\State $\vb*{x}_{near} \gets nearestNeighbor(\vb*{x}_{rand}, T)$ \Comment{(\ref{eq3.2})}
			\State $\vb*{x}_{new} \gets steerOvershoot(\vb*{x}_{rand}, \vb*{x}_{near}, \epsilon, \alpha)$ \Comment{(\ref{eq3.3})}
			\State $g_{new} \gets explorationGain(M, T, \vb*{x}_{new})$ \Comment{(\ref{eq3.4})}
			\State $T \gets (\vb*{x}_{new}, g_{new})$
			\State $N_{T} \gets N_{T} + 1$
			\If {$g_{new} > g_{best}$}
			    \State $\vb*{x}_{best} \gets \vb*{x}_{new}$
			    \State $g_{best} = g_{new}$
			\EndIf
		\EndWhile
		\State $\beta \gets extractNextBestView(T)$
		\State \Return $\beta$
	\end{algorithmic}
\end{algorithm}%

In each loop, the $randomModel$ uniform distribution function randomly selects a point $\vb*{x}_{rand}$ on the map, and the $nearestNeighbor$ function returns the $\vb*{x}_{near}$ vertex of the $T$ tree closest to the point $\vb*{x}_{rand}$, as described in (\ref{eq3.2}).%
\begin{equation}
\vb*{x}_{near} \gets \underset{\forall \vb*{x} \in T}{\arg\min} \|\vb*{x}-\vb*{x}_{rand}\|
\label{eq3.2}
\end{equation}%

The state $\vb*{x}_{new}$, located between $\vb*{x}_{rand}$ and $\vb*{x}_{near}$ on the map, is determined by the $steerOvershoot$ function, ensuring that $|\vb*{x}_{new}-\vb*{x}_{rand}|$ is minimized, $|\vb*{x}_{new}-\vb*{x}_{near}| \leq \epsilon$, and no obstacles exist in the space between $\vb*{x}_{new} + \alpha$ and $\vb*{x}_{near}$, as defined in (\ref{eq3.3}).%
\begin{equation}
\vb*{x}_{new}= \begin{cases}\vb*{x}_{rand}, & \text { if } \|\vb*{x}_{rand}-\vb*{x}_{near}\| \leq \epsilon \\
\vb*{x}_{near}, & \text { if } P_o\left(\vb*{x}_{near}+\alpha, \mathbf{m}\right) \geq 0.5 \\
\vb*{x}_{near}+\epsilon, & \text { otherwise }\end{cases}
\label{eq3.3}
\end{equation}%

The tree $T$ is expanded by adding $\vb*{x}_{new}$ as a new vertex, and an edge is formed by connecting $\vb*{x}_{new}$ and $\vb*{x}_{near}$.

The Sampling-based Method uses the RRT algorithm to explore potential destinations for the robot without extending beyond the observed space. Only points $\vb*{x}_{rand}$ within the known region of the space are sampled. The evaluation function $explorationGain$ is employed to select the optimal nodes of the tree $T$, calculated using (\ref{eq3.4}).%
\begin{equation}
g_{k} = g_{k-1} + G\left(M, \mathbf{m}_{k}\right)  e^{-\lambda \|\vb*{x}_{k}-\vb*{x}_{k-1}\|}
\label{eq3.4}
\end{equation}%

Given that $\vb*{x}_k$ is the node under consideration, $\vb*{x}_{k-1}$ can be obtained through the nearest node of $T$. The value $\lambda$ represents the weight of the distance cost. The function $G\left(M, \mathbf{m}_k\right)$ returns the gain of $\mathbf{m}_k=H_{M} \vb*{x}_k$, referring to their $n$ surrounding cells with radius $r^{gain}_{max}$, weighted by (\ref{eq3.5}), and $\gamma$ is the weight of occupancy probability cost. Occupancy probability $p(\mathbf{m}^i_k)$ for each cell $\mathbf{m}^i_k$ is calculated using (\ref{eq3.6}).%
\begin{equation}
\begin{aligned}
    G\left(M, \mathbf{m}_k\right)=\sum_{i=0}^n e^{-\gamma (1 - 2p\left(\mathbf{m}_k^i\right))}
\end{aligned}
\label{eq3.5}
\end{equation}
\begin{equation}
    \mathbf{m}^i_k \sim U(0, 2r^{gain}_{max}) + m_k
    \label{eq3.6}
\end{equation}%

Eventually, the point $\vb*{x}_{new}$ is considered the current optimal destination of the search tree if its value $g_{new}$ is greater than the previous value of $g_{best}$ of the search tree. Once the loop concludes, the $extractNextBestView$ function returns the first node of branch $T$.

\section{Experiments And Results}
In the evaluation phase, we present a system for autonomous exploration using a mobile robot equipped with a differential drive and a laser sensor. The modified proposed method was assessed through simulated two-dimensional experiments and implemented in a realistic environment. It was compared to the RH-NBV based on \cite{b16}, adapted for a 2D grid occupancy map, and the Sampling-based Frontier Detection method based on \cite{b15}. Note that our proposed technique, the RH-NBV, and the Frontier method utilized the same RRT methodology with identical parameters for constructing RRTs.

\subsection{System Overview}
Our study, referring to \cite{b15}, developed a system for autonomous exploration using an indoor mobile robot with a differential drive for two-dimensional environments. The robot platform includes an Nvidia AGX Xavier embedded computer and a Hokuyo UST-05LX laser sensor. The laser scanner is mounted to scan the environment around the robot, covering a 240-degree front view with 720 sampling points and a maximum range of $8m$.

\begin{figure}[htbp]
    \centering
    \includegraphics[width=.9\columnwidth]{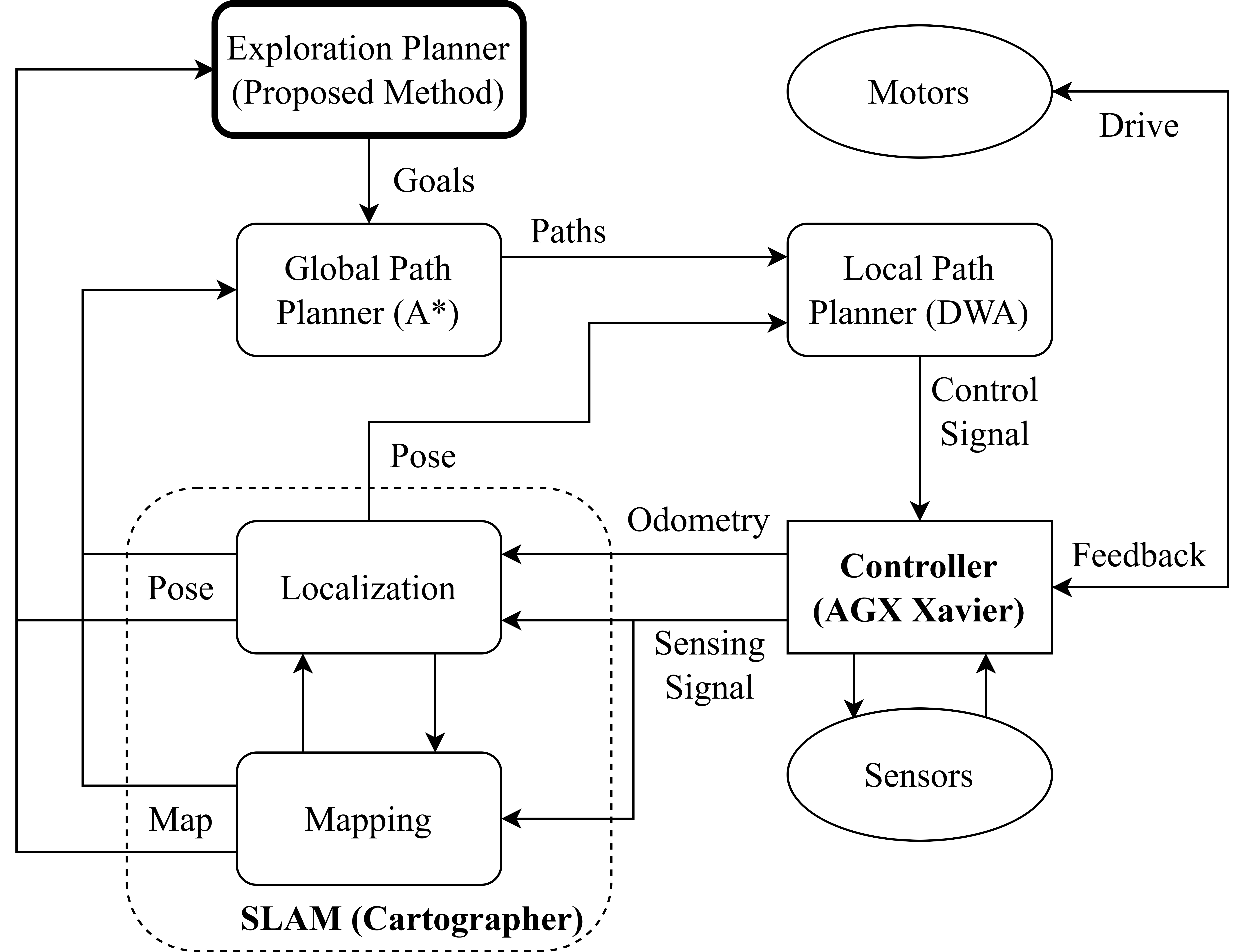}
    \caption{Overview of the experiment system framework. The main modules and their connections are shown in the system diagram.}
    \label{fig:system}
\end{figure}

The proposed method and experimental system are implemented as a Robot Operating System package and tested in our customized environments. The system comprises four components: Cartographer SLAM, A* Global Planning, DWA Local Planning, and our proposed approach, as depicted in Fig. \ref{fig:system}.

\subsection{Scenarios}
This section presents the results of our experiments in both the simulated maze and the realistic environment using identical parameters. The grid occupancy map resolution was set to $r = 0.05m$. The robot traveled with a maximum linear velocity and angular velocity of $v_{max} = 0.3 m/s$ and $\psi_{max} = 1.0 rad/s$, respectively. Some initial RRT and gain function parameters were manually selected for our suggested method, the RH-NBV, and the frontier method, declared by Table \ref{table:params}.%
\begin{table}[htbp]
    \renewcommand{\arraystretch}{1.3}
    \centering
    \caption{Exploration Scenario Parameters}
    \label{table:params}
    % \normalsize
    \begin{tabular}{L{.25\columnwidth} c | L{.25\columnwidth} c}
        \toprule
        \bfseries Parameter & \bfseries Value & \bfseries Parameter & \bfseries Value\\
        \midrule
        Map resolution $r$ & $0.05m$ & $d^{sensor}_{max}$ & $8m$\\
        $v_{max}$ & $0.3 m/s$ & $\psi_{max}$ & $1.0 rad/s$\\
        $N_{max}$ & $15$ & $\lambda$ & $0.5$\\
        $\epsilon$ & $1.5$ & $\alpha$ & $0.3$\\
        $\gamma$ & $4$ & $r^{gain}_{max}$ & $0.3m$\\
        \bottomrule
    \end{tabular}
\end{table}

\subsection{Maze Simulated Environment}
Simulations were conducted in the $20 \times 20m^2$ maze environment. We develop an evaluation of the RH-NBV, Frontier, and our proposed methods by recording the essentials to cover specific average explored spaces after $20$ iterations executing in $900$ seconds. The primary evaluation criteria include the mean and standard deviation of distance, execution time, computation time, and average speed. Additionally, we analyze the success iterations, representing the number of times the robot covered 120 $m^2$, 240 $m^2$, and 360 $m^2$ of the map.

As shown in Table \ref{table:stats}. Our proposed method consistently outperforms the other methods regarding distance traveled, execution time, computation time, and average speed across all coverage levels. Especially at 360 $m^2$ ($90\%$) coverage, the RH-NBV method did not achieve any successful iterations, while our approach achieved most times. This demonstrates that the customized NBV is more effective in covering larger map areas.%
\begin{table*}[htbp]
    \renewcommand{\arraystretch}{1.3}
    \setlength{\tabcolsep}{3pt}
    \centering
    \caption{Statistics of Our Robot Exploration in Maze Simulation Environments}
    \label{table:stats}
    % \normalsize
    \begin{tabular}{L{.08\textwidth} C{.15\textwidth} C{.09\textwidth} C{.15\textwidth} C{.15\textwidth} C{.15\textwidth} C{.09\textwidth}}
        \toprule
        \bfseries Methods & \bfseries Covered Areas & \bfseries Successful Iterations & \bfseries Path Length & \bfseries Execution Time & \bfseries Computation Time & \bfseries Average Speed\\
        \midrule
        \multirow{3}{*}{\bfseries RH-NBV} & $120m^2$ ($30\%$) & $20$ & $25.36 \pm 11.18m$ & $150.42 \pm 80.40s$ & $32.85 \pm 25.71s$ & $0.17m/s$\\
        & $240m^2$ ($60\%$) & $14$ & $106.37 \pm 16.34m$ & $577.75 \pm 105.72s$ & $98.68 \pm 71.75s$ & $0.18m/s$\\
        & $360m^2$ ($90\%$) & $0$ & $NaN$ & $NaN$ & $NaN$ & $NaN$\\
        \midrule
        \multirow{3}{*}{\bfseries Frontier} & $120m^2$ ($30\%$) & $19$ & $27.04 \pm 6.73m$ & $149.63 \pm 59.27s$ & $7.08 \pm 21.32s$ & $0.18m/s$\\
        & $240m^2$ ($60\%$) & $16$ & $71.73 \pm 9.92m$ & $416.62 \pm 135.49s$ & $25.46 \pm 64.66s$ & $0.17m/s$\\
        & $360m^2$ ($90\%$) & $2$ & $114.23 \pm 8.50m$ & $615.57 \pm 121.89s$ & $17.64 \pm 6.77s$ & $0.19m/s$\\
        \midrule
        \multirow{3}{*}{\bfseries Ours} & $120m^2$ ($30\%$) & $20$ & $19.16 \pm 4.19m$ & $93.16 \pm 28.96s$ & $3.50 \pm 0.68s$ & $0.2m/s$\\
        & $240m^2$ ($60\%$) & $20$ & $85.57 \pm 16.45m$ & $401.97 \pm 94.11s$ & $16.35 \pm 5.11s$ & $0.21m/s$\\
        & $360m^2$ ($90\%$) & $16$ & $153.21 \pm 26.62m$ & $716.95 \pm 127.12s$ & $27.78 \pm 9.08s$ & $0.21m/s$\\
        \bottomrule
    \end{tabular}
\end{table*}%

\begin{figure*}[htbp]
    \includegraphics[height=.29\textwidth]{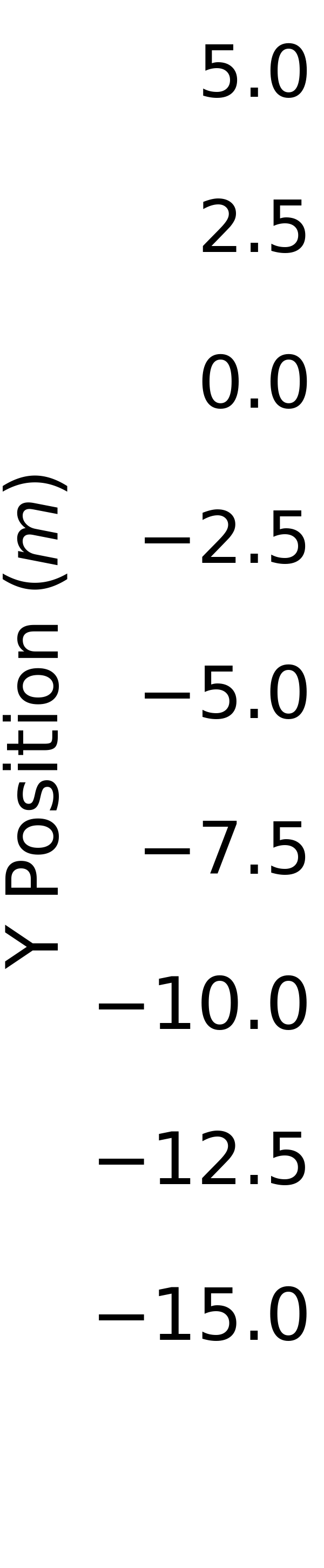}
    \subfloat[RH-NBV Method]{\includegraphics[height=.29\textwidth]{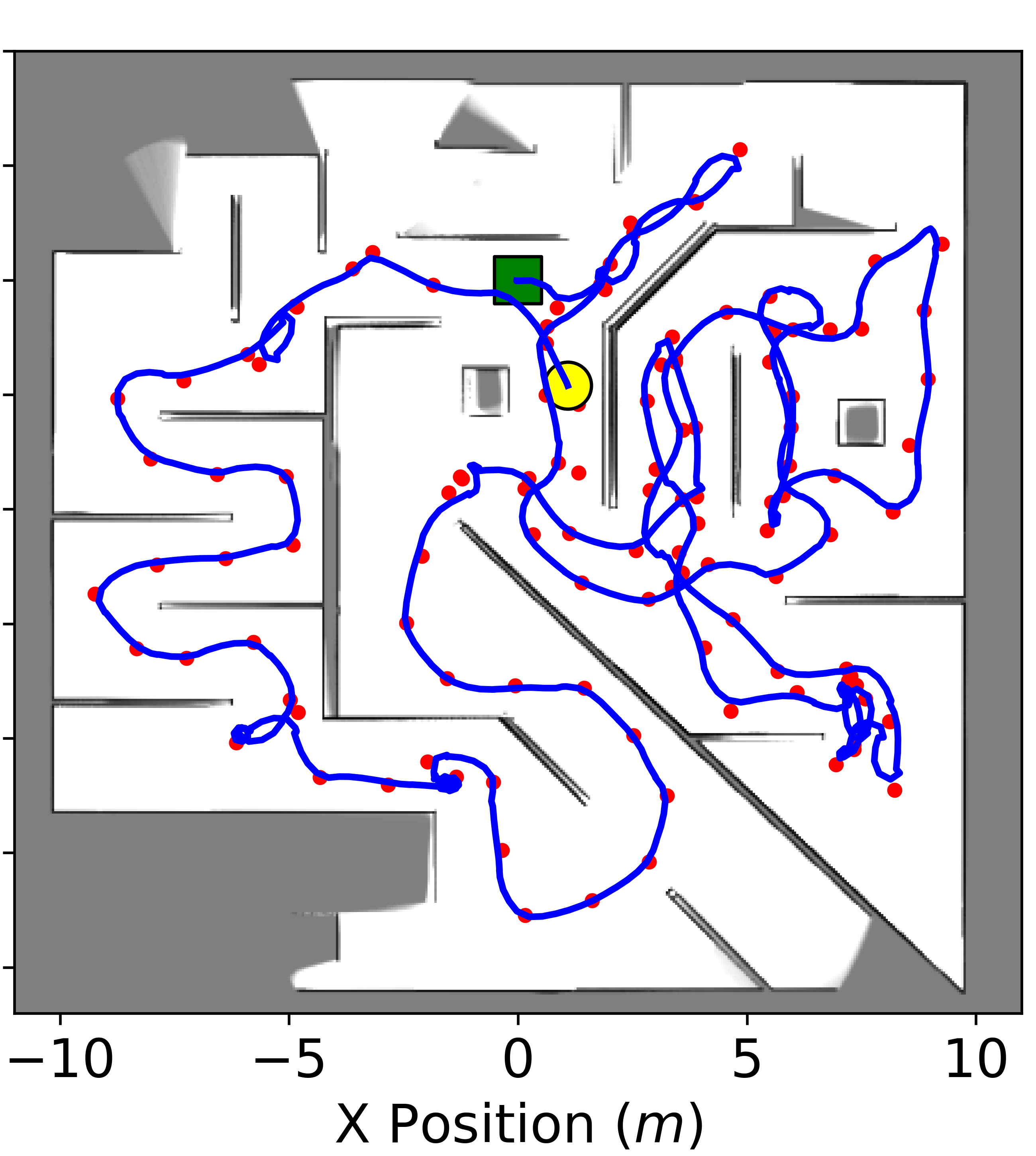}
    \label{fig:map-nbv}}
    \subfloat[Frontier Method]{\includegraphics[height=.29\textwidth]{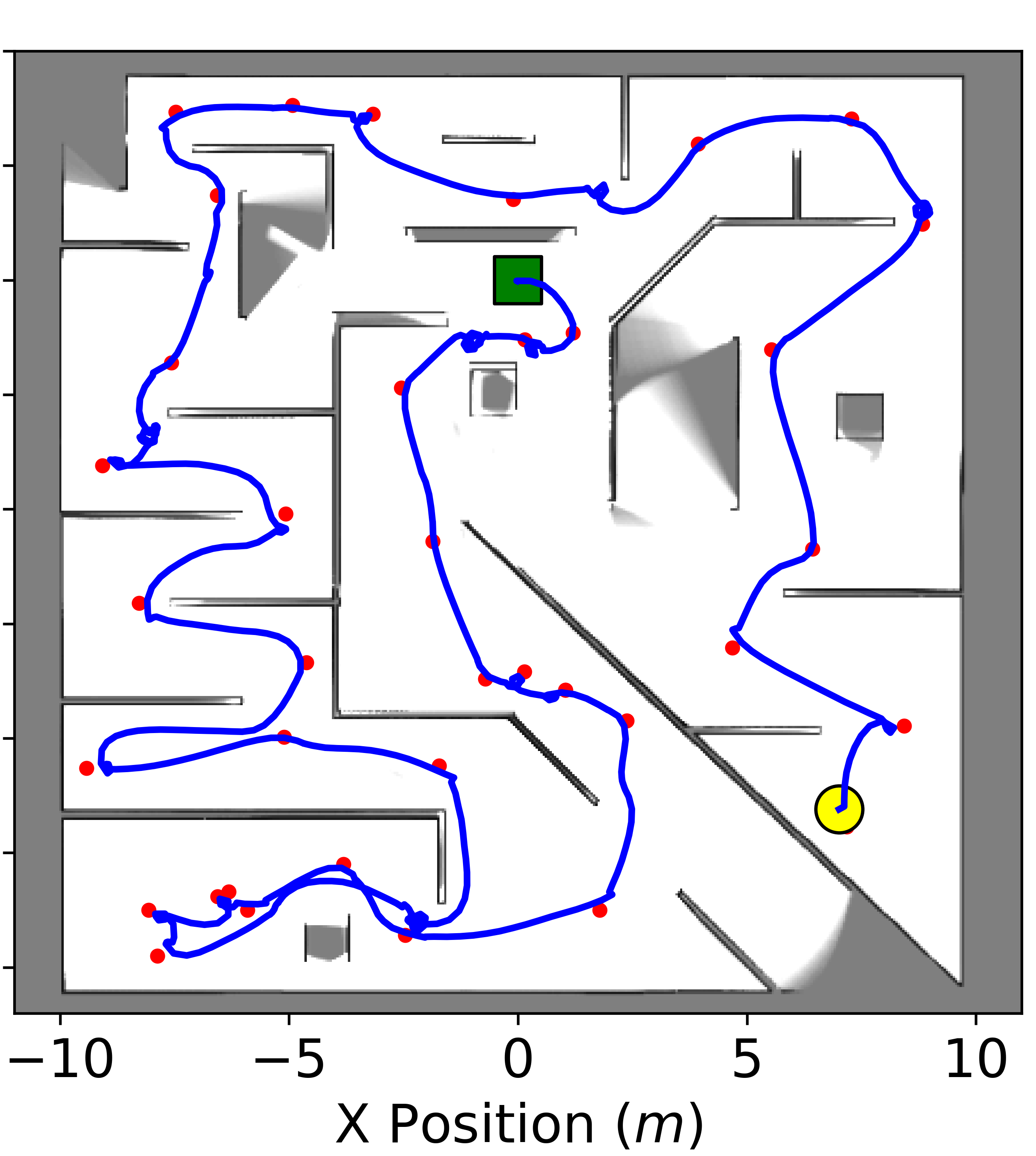}
    \label{fig:map-fd}}
    \subfloat[Our Method]{\includegraphics[height=.29\textwidth]{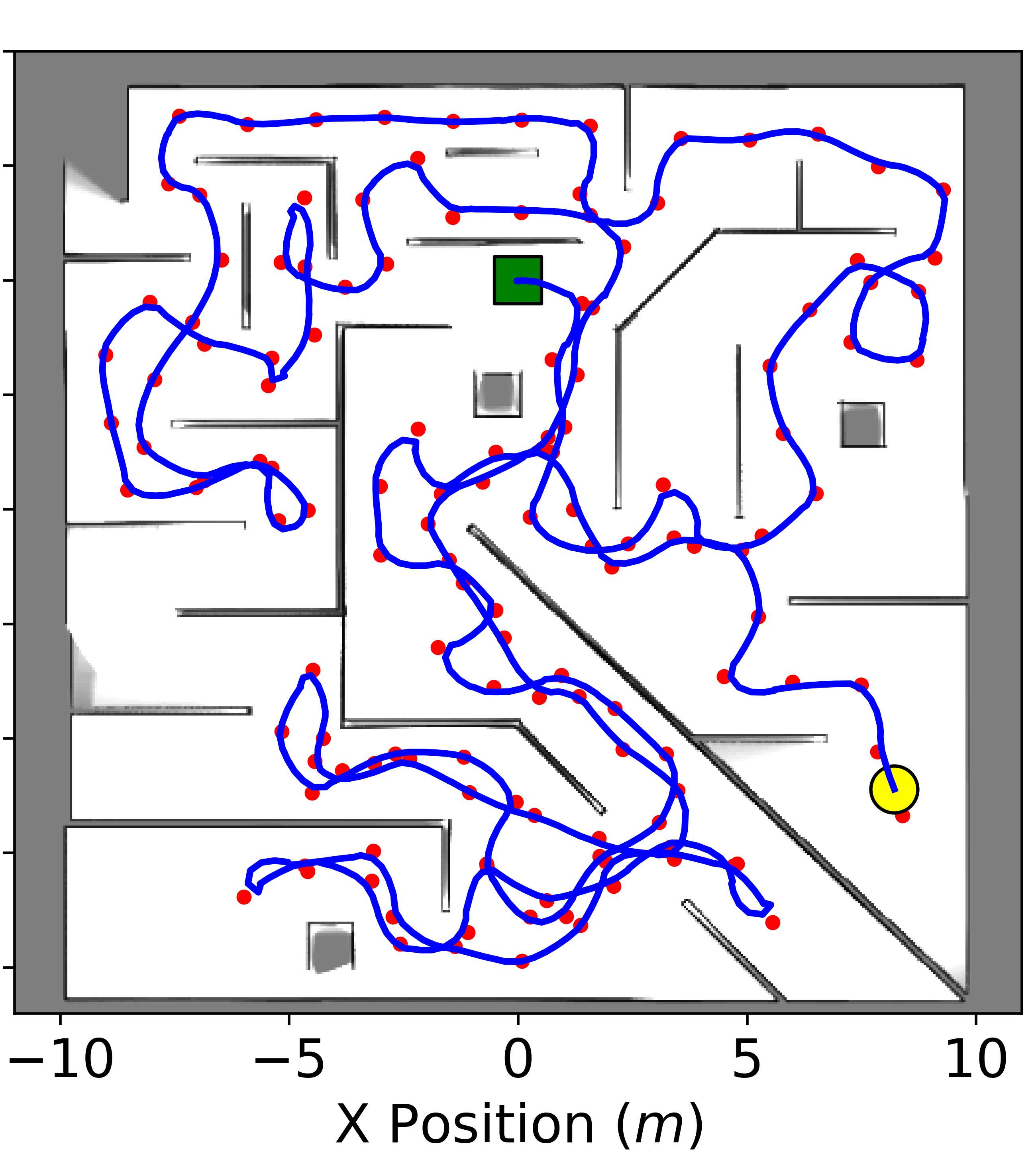}
    \label{fig:map-nbvg}}
    \includegraphics[width=0.95in]{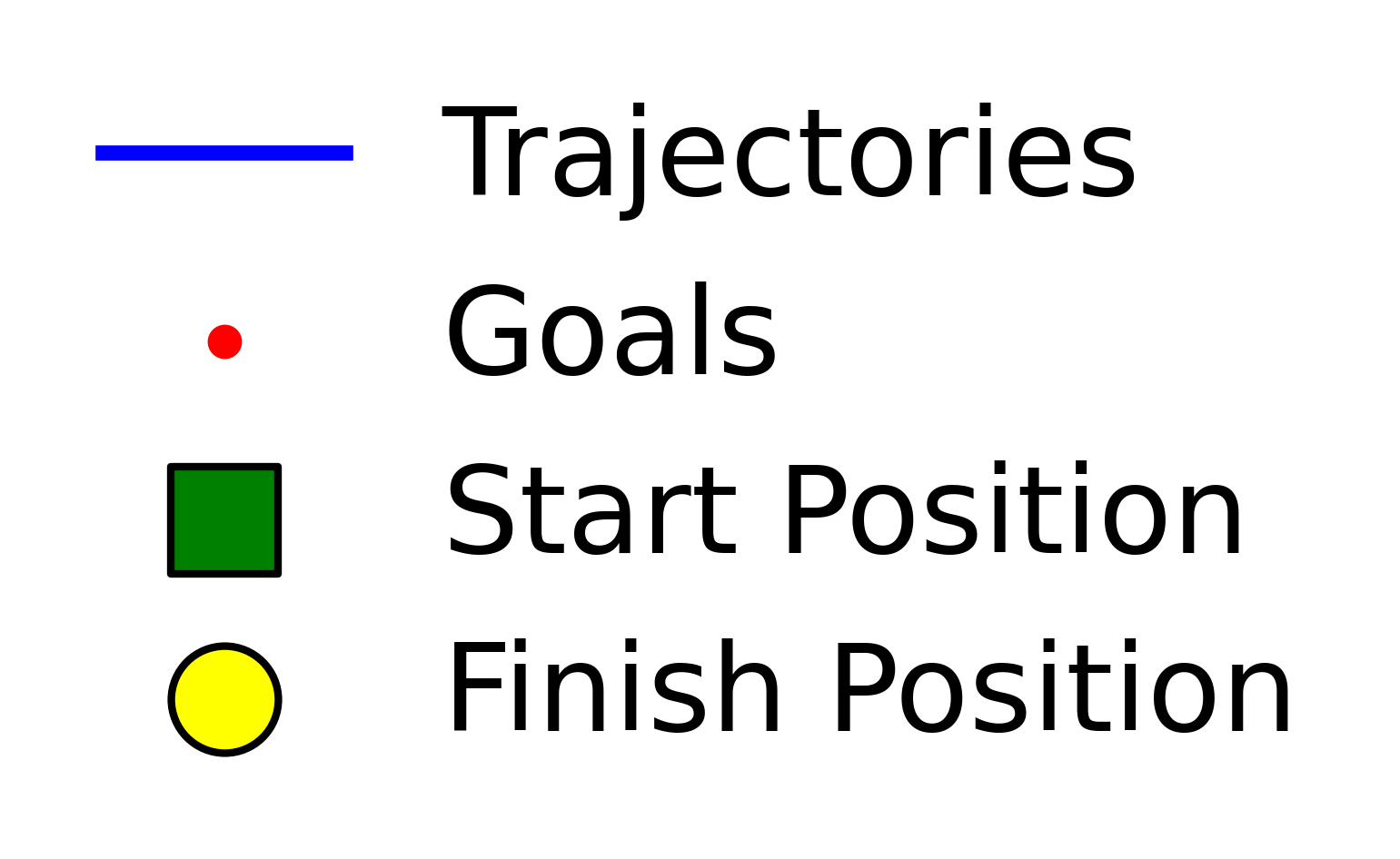}
    \caption{Top-down view images represent maps covered by three methods. The blue lines show the path. The red points show the NBV goals. A green marker is the start position of the robot, and a yellow marker is the robot's position after $900s$ executing each approach.}
    \label{fig:map}
\end{figure*}

Fig. \ref{fig:map} depicts top-down view images of the iteration result in the highest covered areas after $900s$. Our method successfully solved the case, and the entire environment's target exploration ($99\%$) was completed. The frontier method could not see the goal because it was too far and stuck the robot in a harmful stage. In this case, it faced too close obstacles and could not make new motion decisions. With RH-NBV, the deteriorating situation due to immoral actions and uncertain goals makes the robot stuck in small, confined areas. Still, it cannot replan to escape this stage. The modified NBV approach enables the robot to avoid local minima better than the RH-NBV method and be more aware of its surroundings than the Frontier method, resulting in minimized localization errors.%

\begin{figure}[htbp]
    \centering
    \includegraphics[width=.85\columnwidth]{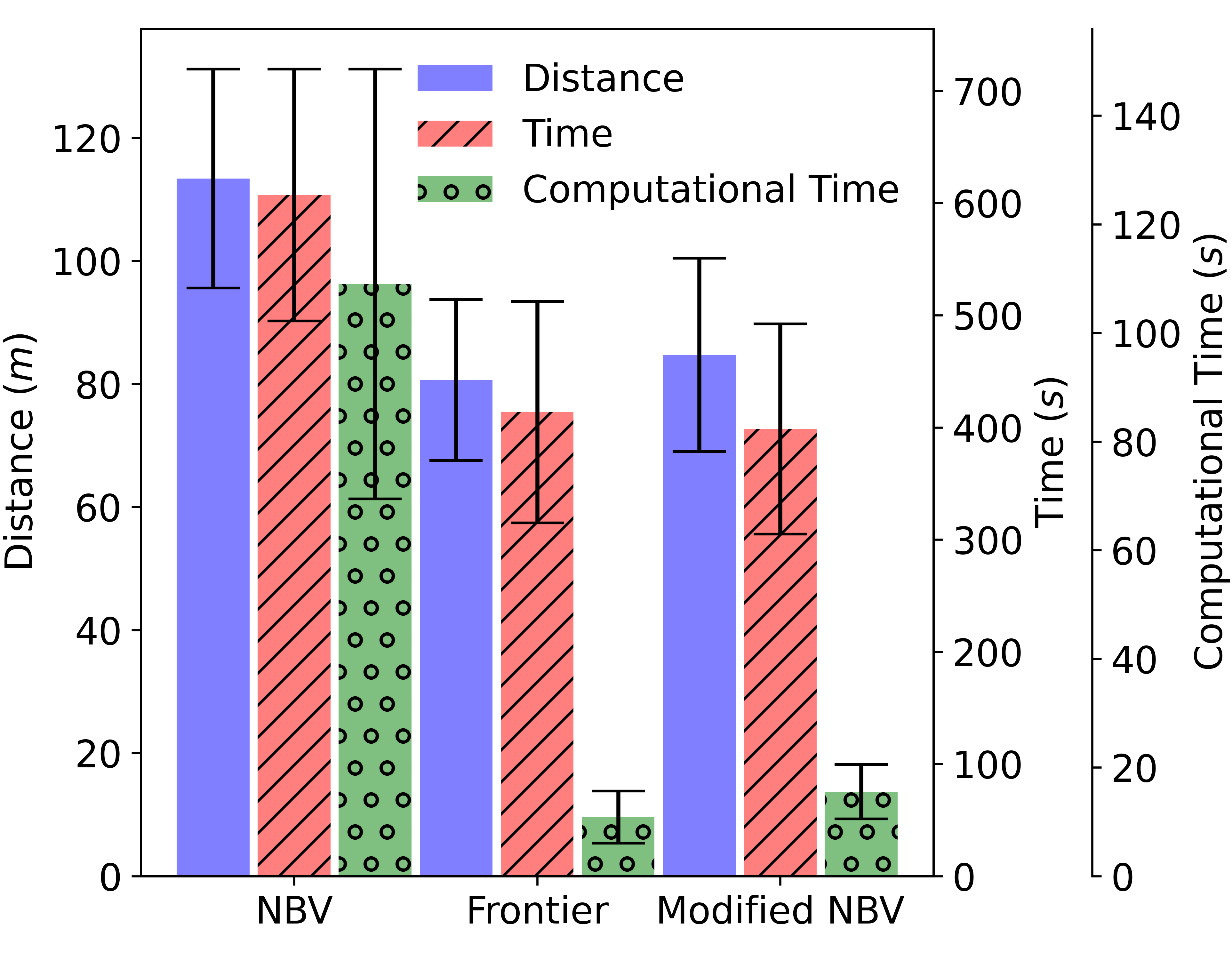}
    \caption{Comparision of travel distance, travel time, and computation time needed to drive the robot to cover $70\%$ area of Maze Environment with different approaches.}
    \label{fig:result-time}
\end{figure}

The performance of the three methods was also evaluated based on the mean and standard deviation of the time, computational time, and distance traveled to reach $70\%$ of the explored area after $20$ runs. Depicted by Fig. \ref{fig:result-time}, the modified NBV method indicates better time and space needed to execute, showcasing efficient exploration performance and precise map traversing in less time and length than the RH-NBV method.%

\subsection{Real Environment}
\begin{figure}[htbp]
    \centering
    \subfloat[]{\includegraphics[width=.39\columnwidth]{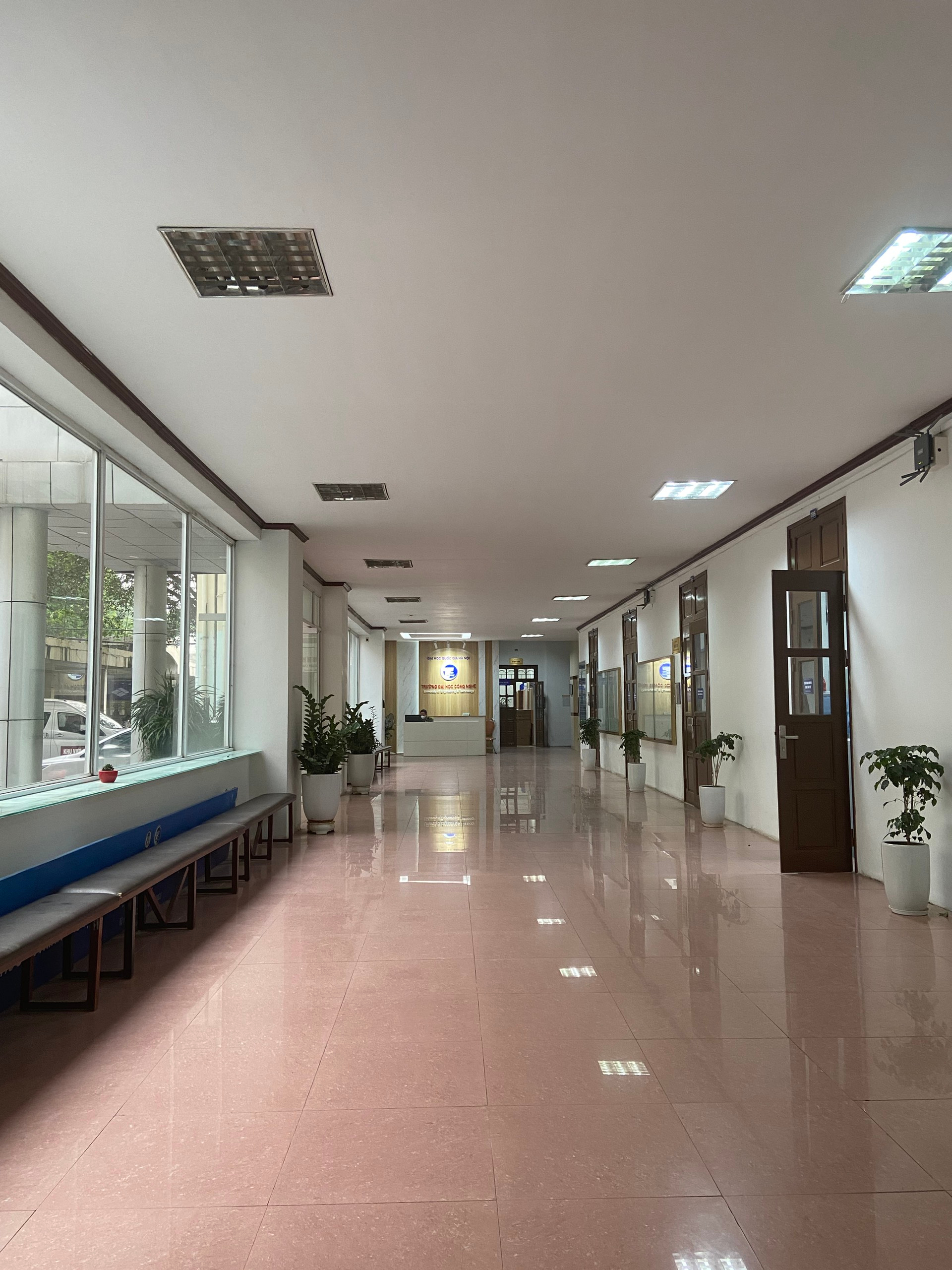}}
    \hspace{1em}
    \subfloat[]{\includegraphics[width=.39\columnwidth]{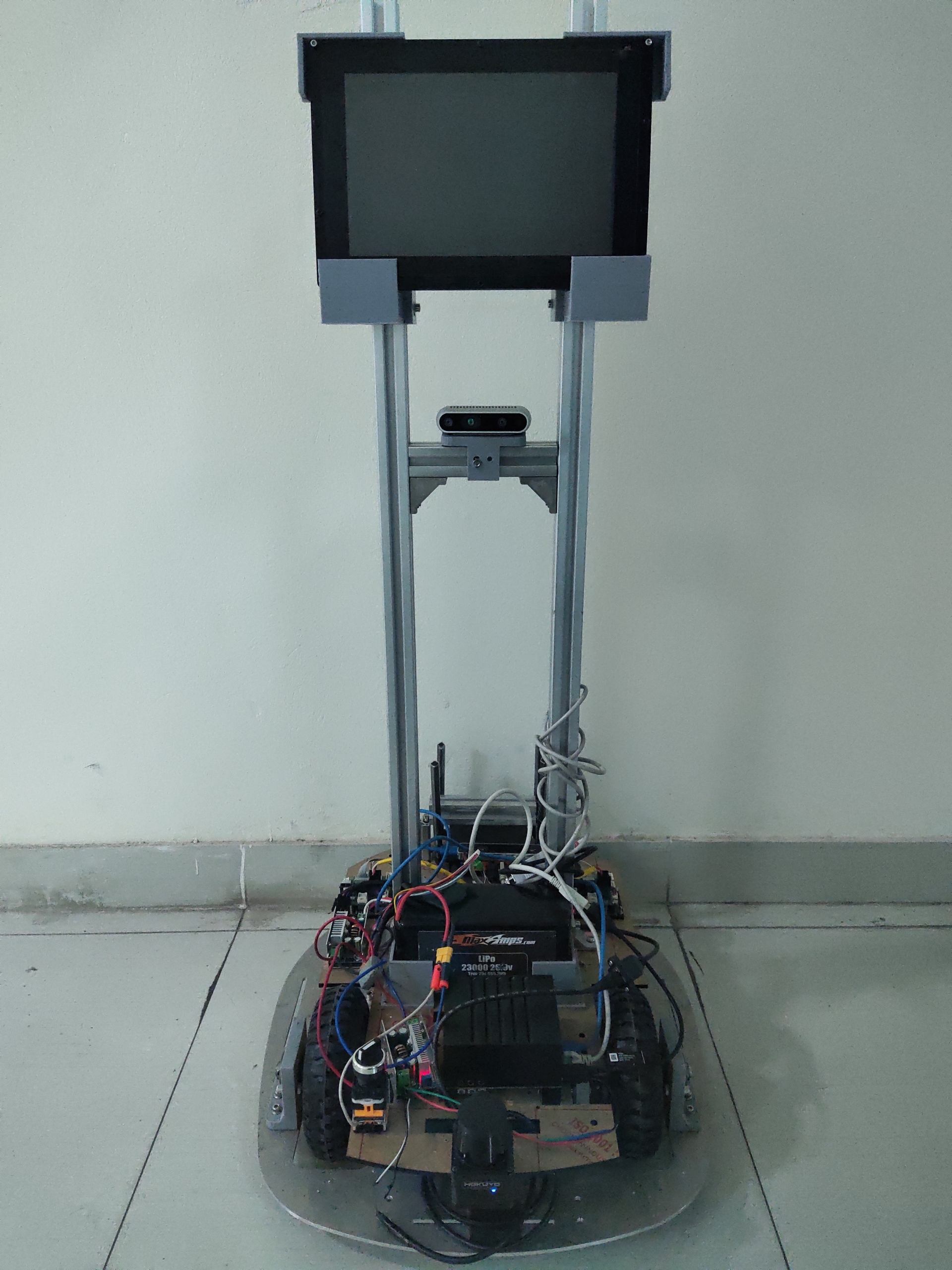}}
    \hfill
    \subfloat[]{\includegraphics[width=.72\columnwidth]{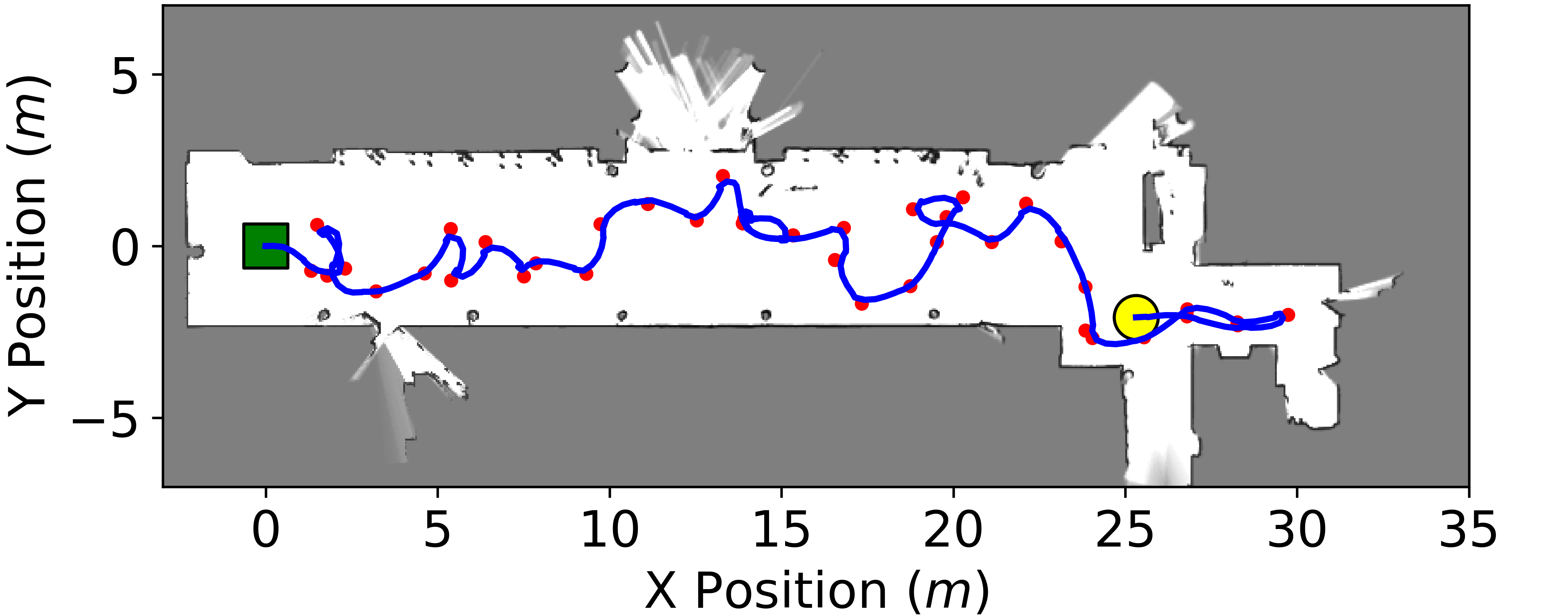}}
    \subfloat{\includegraphics[width=.28\columnwidth]{figures/Maze/legend.png}}
    \caption{Top-down map of our robot exploring the real indoor environment. (a) Real environment. (b) Robot platform. (c) Corresponding occupancy map.}
    \label{fig:map-real}
\end{figure}

In this subsection, robot experiments are conducted in a natural indoor environment to evaluate the stability of our system using the proposed method. We aimed to determine whether the robot could easily map the entire environment. Fig. \ref{fig:map-real} shows a top-down map view after scanning most of the real indoor environment space.%

The robot managed to cover $203.54m^2$ ($97\%$ of the environment) after traveling for $292.93s$, with a distance of $58.44m$, and using $7.54s$ of computation time. These results demonstrate the effectiveness of our proposed Customized NBV method in a real-world setting, highlighting its robustness and applicability for efficient exploration tasks.

\subsection{Discussion}
One of the critical factors contributing to the success of our modified NBV approach is its ability to avoid local minima more effectively than the RH-NBV method. This is primarily achieved through integrating the modified utility metric and boundaries-aware conditions, which enables the robot to make informed decisions about its next destination based on the current state of the environment. In addition, our modified NBV method maintains an uncertain awareness of the map by using additional gain summed from occupancy probability, allowing it to adapt its exploration strategy more effectively in response to environmental changes. This awareness resulted in minimized localization errors and ensured more efficient exploration. Another notable aspect of our modified NBV approach is sampling random map cells for computing gain instead of making clustering or counting unmapped voxels, thus eliminating high computational costs.

\section{Conclusion}
In this study, we presented a modified Next Best View (NBV) approach for autonomous exploration using a mobile robot in two-dimensional environments. Our proposed method combines the benefits of the Frontier approach with an innovative exploration gain function to improve the robot's exploration efficiency and adaptability to its surroundings. The experiments conducted in a simulated maze and a realistic environment demonstrated that our modified NBV method consistently outperforms the Receding Horizon NBV (RH-NBV) and Frontier methods regarding the explored area, time efficiency, and exploration consistency. Comparing our suggested planner to state-of-the-art autonomous sampling-based exploration planners such as RH-NBV and Frontier demonstrates that the proposed algorithm is applicable and can be further refined for specific applications.

Evaluations of the proposed approach in an actual environment using a self-developed mobile robot are in progress. In the future, by using LIDARs and cameras, we intend to construct a comprehensive strategy that can be used in 2D and 3D situations. The scalability of our modified NBV approach to multi-robot systems should be considered. The development of a collaborative exploration strategy, where multiple robots work together to explore the environment, could significantly improve the efficiency and coverage of the exploration process.

\bibliographystyle{IEEEtran}
\bibliography{IEEEabrv, references}

\end{document}